\documentclass[a4paper,twoside]{article}

\usepackage{epsfig}
\usepackage{subcaption}
\usepackage{calc}
\usepackage{amssymb}
\usepackage{amstext}
\usepackage{amsmath}
\usepackage{amsthm}
\usepackage{multicol}
\usepackage{pslatex}
\usepackage{apalike}
\usepackage{algorithm2e}
\usepackage[bottom]{footmisc}
\usepackage{flushend}
\usepackage[hyphens]{url}
\usepackage{SCITEPRESS}     

\begin{document}

\title{Interpretable Text Classification Applied to the Detection of LLM-generated Creative Writing}

\author{
\authorname{Minerva Suvanto\sup{1}\orcidAuthor{0009-0003-1751-151X}, 
Andrea McGlinchey\sup{2}\orcidAuthor{0009-0001-8436-237X}, 
Mattias Wahde\sup{1}\orcidAuthor{0000-0001-6679-637X} and 
Peter J Barclay\sup{2}\orcidAuthor{0009-0002-7369-232X}}
\affiliation{\sup{1}Chalmers University of Technology, Gothenburg, Sweden}
\affiliation{\sup{2}Edinburgh Napier University, Edinburgh, UK}
\email{\{minerva.suvanto, mattias.wahde\}@chalmers.se, andrea.mcglinchey@lumerate.com, p.barclay@napier.ac.uk}
}


\abstract{We consider the problem of distinguishing human-written creative fiction (excerpts from novels) from similar 
text generated by an LLM. Our results show that, while human observers perform poorly (near chance levels) on this binary classification task, a variety of machine-learning models achieve accuracy in the range 0.93 - 0.98 over a previously unseen test set, even using only short samples and single-token (unigram) features. We therefore employ an inherently interpretable (linear) classifier (with a test accuracy of 0.98), in order to elucidate the underlying reasons for this high accuracy. In our analysis, we identify specific unigram features indicative of LLM-generated text, one of the most important being that the LLM
tends to use a larger variety of synonyms,
thereby skewing the
probability distributions in a manner that is easy to detect for a machine learning classifier, yet very difficult
for a human observer. Four additional explanation categories were also identified, namely, temporal drift, Americanisms, foreign language usage, and colloquialisms.
As identification of the AI-generated text depends on a constellation of such features, the classification appears robust, and therefore not easy to circumvent by malicious actors intent on misrepresenting AI-generated text as human work.}

\onecolumn \maketitle \normalsize \setcounter{footnote}{0} \vfill

\section{\uppercase{Introduction}}
\label{sec:introduction}
Text generated via large language models (LLMs), integrated in chatbot applications such as ChatGPT or Gemini, appears with
increasing frequency in a variety of contexts; for example, in academic writing and news articles, as well as in social media content. 

The ease with which such text can be generated has lead to concerns in areas such as academic
plagiarism, spurious product reviews, and misleading political or social commentary 
(\textit{fake news}), as malicious actors can now create misleading content with very little effort. Such concerns have
led to the development of various AI-detection tools to distinguish text written by humans 
from that generated by LLMs. Despite the threat to the livelihood of authors, less attention
has been given to the issue of \textit{sham books}, such as genre novels generated by LLMs
but presented as being works of human creativity. Reliable identification of 
machine-written texts
in the domain of creative fiction would be useful to editors and publishers to
avoid such misrepresentation.

Two earlier papers,~\cite{mcglinchey25} and~\cite{verma_ghostbuster_2023}, investigated the performance of classification methods applied to the problem of distinguishing human-written creative text (such as novels) from LLM-generated texts with similar content. One of the main findings, which is further strengthened by the results obtained in this paper, is that while humans struggle to accurately classify such texts, machine-learning (ML) models, even linear ones, do so with ease. These findings thus raise an interesting research question, which is the main focus of this paper, namely:

\textit{What aspects of the texts make it possible even for simple ML text classifiers to obtain
very high classification accuracy?}

A corollary to this question is whether knowing the basis of such successful classification could perhaps even \textit{assist} bad actors in making more deceptive AI-generated text.

The utility of reliably explaining the classification results goes beyond gaining general insights about the differences between the two text categories: While AI-detectors are becoming a growing area of interest, there are, from our viewpoint, problems concerning their reliability and trustworthiness. In particular, detection tools that are based on black-box methods offer either no reasoning for a given classification result, or do so using methods that have often been found to be unreliable or even contradictory. This lack of reliable explanations can weaken the trust in such systems. Therefore, in this paper, we focus primarily on a linear classification method devised with interpretability in mind~\cite{wahde_interpretable_2024}, in order to investigate why ML models are so successful in distinguishing texts belonging to either of the two categories mentioned above.

The structure of the paper is as follows: In the next section, we provide a general description of the problem at hand, as well as some results from related studies. We also emphasise the importance of interpretability. Next, in Section~\ref{sect:methodology}, we describe  
how the data were generated as well as the various ML classifiers used. Our results are presented in Section~\ref{sec:results}, which is followed by a thorough analysis in
Section~\ref{sect:analysis}. Finally, we offer some conclusions in Section~\ref{sec:conclusion}.

\section{\uppercase{Background}}
\label{sec:bground}

\subsection{Text Generation}

Advances in generative AI have enabled companies to integrate LLMs in software programs used in many contexts, thus leading to 
new possibilities for producing textual content, which can now
be generated to a given brief with little time and effort. 
It has become more common to encounter AI-generated reports, product descriptions, reviews, e-mails, speeches, and a multitude of other common text genres.

The models that power modern text generation tools are mainly based on the Transformer architecture~\cite{vaswani2017attention}. These types of models produce text via 
next-token prediction, where, starting from a given text input (a \textit{prompt}), 
the model predicts the most likely output to follow, one token\footnote{Tokens can be words, parts of words, or even individual characters. In the study presented here, the texts are tokenised on the level of individual words (and punctuation symbols, see Section~\ref{sect:classifiers}).} at a time. 
When generating text, previous output tokens are also added to 
the context used by the LLM
for selecting its next token.
Such models are generally trained on enormous amounts of text data, making them able to mimic human writing so well as to produce grammatically correct, coherent text that often appears to be written by a real person.

Replacing human authors with generative AI may seem
to be an option for quickly creating content for various platforms, but doing so has several problematic aspects that should be considered. Some well-studied problems are related to LLMs producing factually incorrect statements and offensive or biased outputs~\cite{dong2023probingexplicitimplicitgender}. Furthermore, as LLMs have been trained on text from existing published works, recent lawsuits have brought to attention that LLM-generated text may be in violation of copyright laws, as they may produce plagiarised 
text~\cite{milmo2025meta,sag2025globalization}. The use of generative AI also threatens the livelihood of human authors, and arguably reduces the quality of the works that are published.

In creative fiction, the focus of this paper, automated text generation can be both
a useful tool for writers, and also a threat to their livelihood.
Tools to assist writers can support the creative writing process and help writers get past writer's block~\cite{yuan2022wordcraft,qin2024charactermeet}. On the other hand, some studies ~\cite{niloy2024chatgpt} have also shown that the use of AI tools has adverse effects on creative writing. One study specifically encourages writers to rewrite AI-generated outputs, which can give a stronger sense of ownership over the produced work~\cite{zhou2024ai}. While most tools studied in the literature seem to encourage using AI as an assistant, it is also possible to create novels completely written by AI~\cite{cabezas2024book}. In some cases, such automatically generated
books have been misrepresented as works of human creativity. 

Reliably detecting whether text was written by a human or generated by an LLM can help to address some of these problems. This task has proven difficult for human evaluators, but automated tools for this purpose show promising results, as shown below.

\subsection{Related Work}

Studies on the detection of artificially generated text have addressed domains such as academic plagiarism, \textit{fake news}, and spurious product reviews \cite{crothers_machine-generated_2023}; 
however, little attention has been given to creative writing, despite rising concern among creative workers as demonstrated by recent authors' protests \cite{creamer2025}.  

Two earlier studies addressed this domain.
The \textit{Ghostbuster} study \cite{verma_ghostbuster_2023} used a sophisticated
multi-stage model, and included a creative writing data set along with other types
of text. However, this data set was not based on published novels, but samples
from a Reddit forum where users write stories based on a writing prompt (story 
outline); the same
writing prompts were then used to generate the AI text, prompting as follows: 
\textit{Write a story in \{length\} words to the prompt: \{prompt\}}.
The \textit{Ghostbuster} approach
achieved a high accuracy (F1 score of 98.4\%) over the creative fiction data set,
but when it was applied to shorter samples of creative text, around 100 tokens
in length, their F1 score dropped to around 85\%.

The only other study to address the area, \textit{AI Detective} \cite{mcglinchey25}
took a simpler approach, using classical ML classifiers from Sci-kit Learn, and
achieved comparable accuracy (F1 between 94.2\% and 100\%, depending on the trial)
using text from published novels. Notably, the \textit{AI Detective} classifiers
performed well even over short samples of around 100 words.

Both these studies demonstrated a remarkable level of accuracy in the identification of AI-generated creative fiction. In this paper, we extend this work by focusing attention on \textit{explaining} the high classification performance for this identification task.

Another recent study~\cite{reinhart2025llms} on the broader topic of general text classification used Biber's linguistic features~\cite{biber1991variation} to classify text as AI-generated or human-written. The study used different categories of text types, including fiction, and also attempted to detect specific LLMs that had written the texts. Inspired by this work, we use the the same feature set in a part of our analysis below. A different study~\cite{mao2024raidar} used a procedure where an LLM was first used to rewrite a text excerpt, then the number of changes that occur during rewriting was measured and used as a basis for detecting AI-generated text. That study found that LLMs change human-written text considerably more than 
they change LLM-generated text, a notion that also aligns with our analysis on LLM-rewritings presented in Section~\ref{sect:analysis}.

\subsection{Interpretability}

Deep neural networks (DNNs), a category that also includes LLMs, are inherently black boxes, meaning that the actual processing underlying their decision-making is not transparent. From the viewpoint of the user, such models take an input and then generate an output, but the manner in which they do so is largely opaque, owing to the highly complex, non-linear model architectures used in neural systems. In many scenarios, understanding the reasons for a model's outputs is essential, and therefore studies on explainable AI and interpretable AI are emerging.

Here, we distinguish between, on the one hand, \textit{explainable AI}, which refers to methods and processes aimed at trying to clarify the decision-making process of a black box (such as a DNN) and, on the other hand, \textit{interpretable AI} that, in our definition, involves inherently interpretable models, for example, rule-based or even linear systems. While much effort has been devoted to attempts at explaining black boxes, the explanations obtained (by methods such as SHAP, LIME, and so on) are often case-specific (applied \textit{a posteriori}), incomplete, unreliable, and sometimes contradictory~\cite{krishna2022disagreement}. Thus, like some other authors~\cite{rudin2019stop}, we prefer to use interpretable models for which the explanations are exact rather than involving some form of linearisation or other forms of simplification, especially in cases like ours where (as shown in Section~\ref{sec:results}) interpretable models already achieve near-perfect accuracy, thus removing any need for the use of more complex models.

In the case of detecting AI-generated text, we argue that the underlying reasons for the classification are crucial in order for such tools to be used reliably as a support for decision-making. By contrast, trusting blindly in a (potentially incorrect) decision made by a black box detection system can, for example, lead to unjustified accusations of AI being used for writing. The detector's results can be rendered more useful with an interpretable method, which provides the user with insights that can be helpful in making a well-informed decision.

However, for the effective detection of automatically generated creative fiction, we face a \textit{paradox of interpretability} -- to have confidence in the results, we need to be able to explain them; but if we can point out just how sham text can be identified, we make it easier for bad actors to create such text in the future. The situation will be worse where one or two key features aid the classification, but if a wide constellation of features are used together, the interpretation will be of less assistance to bad actors.

\section{\uppercase{Methodology}}
\label{sect:methodology}

\subsection{Data Sets}
\label{sect:data}

Following the approach used in~\cite{mcglinchey25}, we have focused on classic detective fiction, 
a genre which (1) is well-known, (2) uses everyday (and not too formal or dated) language, and
(3) has a substantial corpus of material available online. We constructed a data set
named DET12, which is based on text samples from twelve 
published novels.

\begin{table*}[htbp]
\caption{A human-written excerpt (top) and a version (bottom) rewritten by an OpenAI model (GPT-4.1).}
\footnotesize
\centering
\begin{tabular}{| p{2.5cm} | p{12.3cm} |}\hline
\textbf{Text Preparation}   &  \textbf{Text Excerpt} \\
\hline
\textbf{Original Text}   & One was a terrible crockery-smasher, and Miss Whittaker gave her notice, so the other left with her.” “Ah, well! one can have too much of seeing the Crown Derby rollin’ round the floor. Quite. Then it had nothing to do with—it wasn’t on account of any little—” “It wasn’t because they couldn’t get along with the nurse, if you mean that,” said Nurse Philliter, with a smile. “They were very obliging girls, but not very bright.” “Quite. Well, now, is there any little odd, out-of-the-way incident you can think of that might throw light on the thing. There was a visit from a lawyer, I believe, that agitated your patient quite a lot. \\
\hline
\textbf{Model GPT 4.1}   & One of them had a habit of breaking dishes, and Miss Whittaker resigned, so the second girl decided to go too.” “Ah, yes! Too much Crown Derby crashing about can be overwhelming. Indeed. So it wasn’t anything to do with—no sort of—” “It wasn’t a problem with the nurse, if that’s your suggestion,” Nurse Philliter replied cheerfully. “The girls tried to be helpful, though not especially clever.” “Indeed. So, is there any unusual, perhaps overlooked event you recall that might shed some light on what happened? I heard a lawyer called, which greatly unsettled your patient.” \\\hline
\end{tabular}
\label{tab:wording}
\end{table*}

Using copyright-free texts
from Project Gutenberg\footnote{\url{https://www.gutenberg.org/}.}, we selected six popular
detective novels by Agatha Christie\footnote
{``The Murder on the Links'', ``Poirot Investigates'', ``The Man in the Brown Suit'', ``The Mysterious Affair at Styles'', ``The Big Four'' \& ``The Secret Adversary''.} 
and six by Dorothy L.~Sayers\footnote
{``Strong Poison'', ``Unnatural Death'', ``Clouds of Witness'', ``Lord Peter Views the Body'', ``The Unpleasantness at the Bellona Club'', \& ``Whose Body? A Lord Peter Wimsey Novel''.}, two authors writing
in the same genre around the same time. Extraneous matter such as page numbers or chapter headings were removed using a Python script,
then the source texts were chunked into short excerpts of approximately 100 words, 
always cutting on a full-stop\footnote
{Other symbols such as quotes or question marks had been investigated; however, owing to the large amount of direct speech in the novels, they could not be reliably used as breakpoints. We found that prompting the LLM to count spaces was for some reason more reliable than asking it to count words.}
to avoid breaking up sentences. 
A range of 92 to 125 words was used for maximum length, where a length within that range was randomly assigned per section to increase the variability in length in order to reduce potential bias. 
This process yielded 8,068 short samples of human-written text. We note that while the texts are overwhelmingly in English, they do contain quotations and short passages in several other languages (mainly French); we did not see any reason to remove these.

These samples were then sent to an LLM in a random order via the OpenAI
API, with a prompt to rewrite them while preserving the meaning, style, and approximate length, creating 8,068 AI-generated text samples.
The LLM version used for this task was GPT-4.1 and the temperature was set to 0.7. 
The following prompt was used:  

{\small
\textit{You will take the role of an author of crime
novels. A text excerpt will be provided, you
have to review it for the number of space
characters and key details. Create a new 
text excerpt which contains the same key 
details but appears structurally different to 
the original. The new text must have 
approximately the same number of spaces as 
the original. Only return the new text passage. 
Do not include place holders, line breaks or any 
other text except the new passage. 
Text excerpt:''}
}

An example of a human-written text and the corresponding LLM-generated version is
shown in Table~\ref{tab:wording}.

While the mean length of the human-written (Class 0) and LLM-generated
(Class 1) texts was similar, as shown in Table~\ref{tab:data-len}, the
LLM had a preference for generating shorter fragments overall.
Correspondingly, Class 1 had a slightly higher standard deviation, reflecting the 
presence of some longer outliers which pull up the average length
to match the original samples. Note that the averages in Table~\ref{tab:data-len}
also include punctuation symbols, dashes, quotation symbols, and so on, thus explaining why
the averages exceed the maximum length (number of words) stipulated above.

The labelled original and rewritten text samples were then 
split into training (70\%), validation (15\%) and test (15\%) sets which were 
used in the training process for the ML classifier models as described below.

\begin{table}[htbp]
\caption{Length distribution (number of tokens) of text samples. Here, $C$ specifices the
class (0 for human-written, 1 for LLM-generated), $\overline{N}$ is the average number
of tokens of a text sample, whereas $\sigma$ is the corresponding standard deviation.}
 \centering
 \begin{tabular}{|c|c|c|c|c|c|c|}\hline
\textbf{C} & $\overline{N}_{tr}$ & $\sigma_{tr}$ & $\overline{N}_{val}$ & $\sigma_{val}$& $\overline{N}_{tst}$ & $\sigma_{tst}$ \\
\hline
0 & 138.1 & 23.5 & 137.2 & 23.1 & 138.2 & 22.9 \\\hline
1 & 132.6 & 24.4 & 131.2 & 23.9 & 132.4 & 23.7 \\ \hline
\end{tabular}
\label{tab:data-len}
\end{table}

\subsection{Text Classifiers and Features}
\label{sect:classifiers}
Several standard ML classifiers often applied in text classification tasks were used, employing
holdout validation for the training process; see also Section~\ref{sect:mlresults} below.
Specifically, the classifiers used were (i) a linear, explicitly interpretable classifier described in~\cite{wahde_interpretable_2024}, 
(ii) a multinomial Na\"ive Bayes classifier, (iii) a Support-vector machine (SVM)
classifier, (iv) a Logistic regression classifier, and (v) a Random forest classifier.

In all cases, to obtain a fair comparison, each classifier used the same bag-of-words (unigram) feature set, obtained from the training set, as described below.
Despite its simplicity, this feature set, which contained a total of 30,302 features, turned out to be sufficient; see also Section~\ref{sect:mlresults}.
For the linear classifier, unlike the analysis in~\cite{wahde_interpretable_2024}, we 
did \textit{not} include the text length as a feature. Doing so would have helped the ML classifier 
in an unfair way, especially in comparison with human evaluators. For the other classifiers,
we used the implementation in SciKit-Learn, with default settings for Multinomial Na\"ive Bayes,
Logistic regression, and Random forest. For the SVM we used a linear kernel with a penalty
term ($C$) equal to 1.

The feature set was extracted from the training set after preprocessing and tokenising the data set. We applied our own custom method for tokenising the data instead of using the default tokeniser from SciKit-Learn. The inclusive tokenisation method maintains original word casings, and includes most punctuation marks as individual tokens, with the exception of normalising single and double quotations marks and dashes. This was done so that the punctuation marks used within the two text categories were consistent, thus eliminating the inclusion of any obvious indicators of LLM-generated text (see also Section~\ref{sect:artefacts} below). The tokenised data sets are available on Zenodo\footnote{\url{https://zenodo.org/records/17880411}}.

\subsection{Artefacts of the Generation Process}
\label{sect:artefacts}
As shown in Table~\ref{tab:classifier_results} below, the ML classifiers generally obtained a very high accuracy.
Before proceeding, we ensured that the text generation and preparation process did not introduce any artefacts that would be an obvious clue for the classifiers. Systematic differences in layout or punctuation might not be obvious to human evaluators but could bias the classifiers -- for example, using a longer type of dash in the output text than was present in the input (say Unicode U+2013 versus U+2014). 

In order to eliminate such possibilities, before carrying out the final tokenisation described above, we ran some preliminary experiments (i) using the default Scikit-Learn tokeniser, (ii) using a custom-written tokeniser (mentioned above), which standardised the punctuation, and (iii) using a data set where all punctuation and capitalisation had been removed from the texts. In all cases, there was no discernible difference in the classification accuracy. In the end, we chose to use our custom-written tokeniser; see Section~\ref{sect:classifiers}.
We also checked that the distribution of individual letters a--z followed the usual pattern for English text, and was consistent across both classes. These checks gave us confidence that the classification relied on word choice, on which we focus the subsequent analysis.
\section{\uppercase{Results}}
\label{sec:results}

In general, human evaluators are not very successful in classifying text as human-written or LLM-generated~\cite{clark_all_2021,gao2023comparing}.
Similar findings were also made for the specific case of creative fiction~\cite{mcglinchey25}. On the other hand, for
the case considered here, ML classifiers solve this task with ease, as will be shown below.

\subsection{Classification by Human Evaluators}
\label{sect:humanresults}

In~\cite{mcglinchey25}, 19 human evaluators assessed 10 text excerpts (either human-written or LLM-generated), achieving 55\% accuracy. Here, we repeated their experiment, this
time with five (different) text pairs, but with as many as 119 human evaluators.
While only a few evaluators were native English speakers, all evaluators had a
good command of the English language. The evaluators were shown two text excerpts (at the same time),
one human-written and one LLM-generated, from our test set. The excerpts were quite
short, around 4-7 lines each. They were then asked to determine which of
the two excerpts was LLM-generated. For each pair, the participants were given 90 seconds to make
their assessment, and they did not have Internet access during the experiment.
We thus collected a total of $5 \times 119 = 595$ answers. Of those, 297 were correct,
thus resulting in an overall accuracy of $297/595 \approx 0.499$, \textit{i.e.}, almost exactly a random outcome. In other words, it is clear that human evaluators are generally \textit{not} capable of distinguishing
LLM-generated excerpts from human-written ones.

\begin{table}[htbp]
    \caption{Classification accuracies obtained by different ML classifiers over the test set of the DET12 data set; see also
    Section~\ref{sect:mlresults}.}
    \centering
    \begin{tabular}{|l|l|c|}\hline
            \textbf{Classifier} & \textbf{Implementation} & \textbf{Test acc.} \\\hline
        Linear        & Ours         & 0.9814 \\\hline
        Na\"ive Bayes   & SciKit-Learn & 0.9793 \\\hline
        (Linear) SVM           & SciKit-Learn & 0.9694 \\\hline
        Logistic regr.     & SciKit-Learn & 0.9674 \\\hline
        Random Forest & SciKit-Learn & 0.9310 \\\hline
    \end{tabular}
    \label{tab:classifier_results}
\end{table}
\subsection{Classification Using ML Methods}
\label{sect:mlresults}
We then proceeded to classify the text samples using the various ML methods listed in Section~\ref{sect:classifiers}.
In all cases, only unigram features were used, for the simple reason that it turned out to be sufficient: Using
more complex features (such as, for instance, n-grams with $n > 1$, TF-IDF features, or linguistic features such as Biber's feature set
~\cite{biber1991variation}) had only a very minor effect on \textit{classification} accuracy, even though an analysis of linguistic features (presented below) provided some interesting clues for \textit{interpreting} our results. Moreover, focusing on unigram features makes it possible to visualise the difference between the two classes in a very 
clear manner, as illustrated below.

During training of the linear classifier and the SVM classifier, holdout validation was used, such that the performance over the training set was given as feedback to the
respective training algorithm, whereas the validation performance was used to determine when to stop training (selecting the classifier with the best validation performance). For the Na\"ive Bayes, Random forest, and Logistic regression classifiers there was no iterative tuning (hyperparameter tuning was not applied). Here, the training algorithm
is simply executed once, resulting in \textit{one} 
trained classifier; thus the validation set was effectively not needed in those cases.

\begin{figure}
    \centerline{
    \includegraphics[width=\columnwidth]{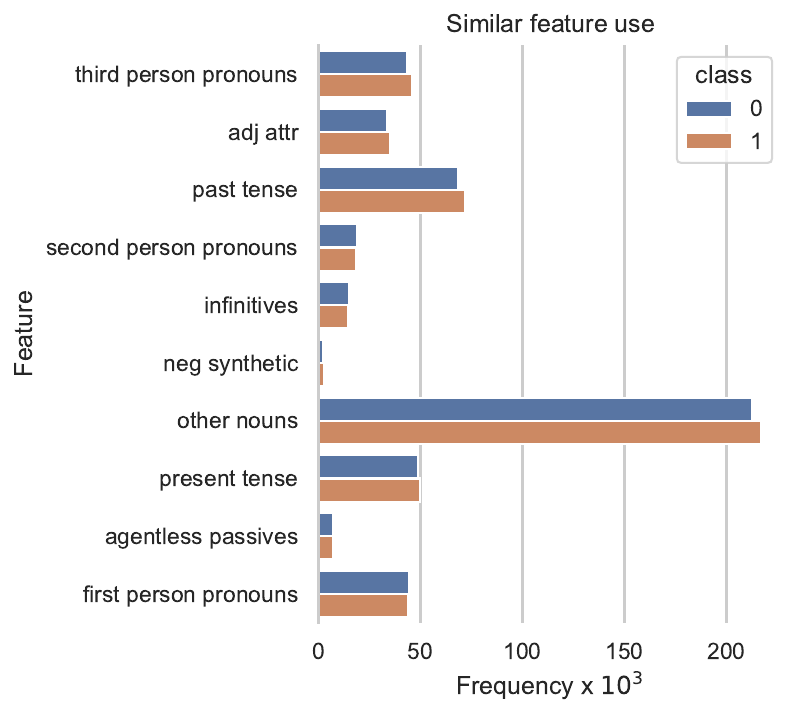}
    }
    \caption{Analysis of linguistic features on our DET12 training data set, showing the 10 features that exhibited the smallest difference between the two classes.
    The horizontal axis measures the (relative) frequency, \textit{i.e.}, the number of
    instances per million tokens.}
    \label{fig:biber_features_similar}
\end{figure}    
    
\begin{figure*}
    \centerline{
    \includegraphics[width=0.49\linewidth]{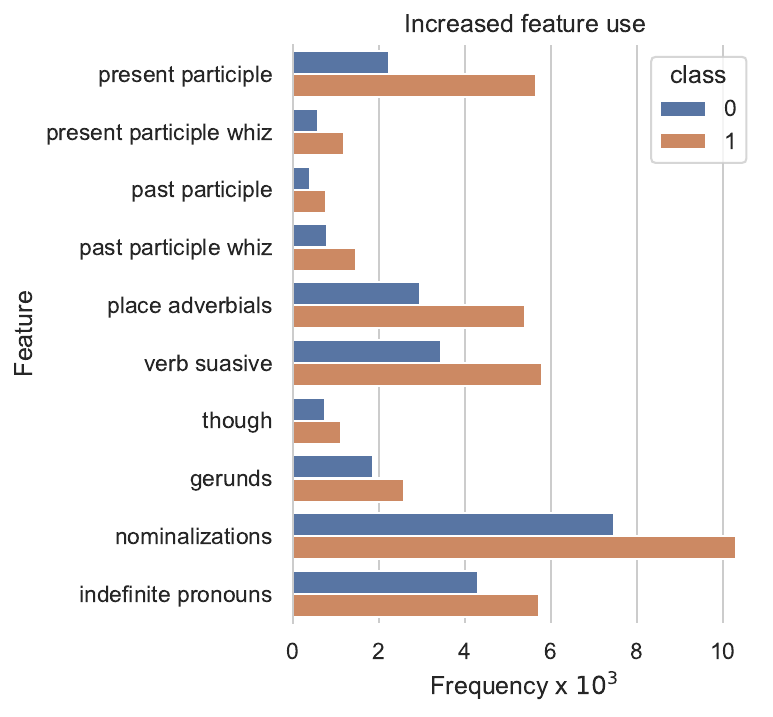}\hspace{0.02\linewidth}
    \includegraphics[width=0.49\linewidth]{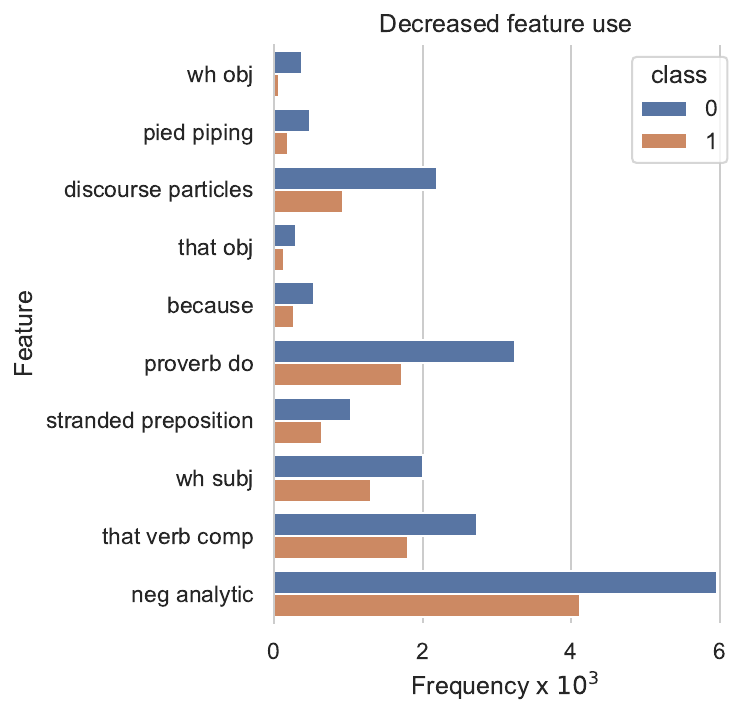}
    }
    \caption{Analysis of linguistic features on our DET12 training data set. The left panel shows the 10 features that had the largest frequency increase in Class 1 (when compared to Class 0), while the right panel shows the 10 features with the largest decrease in Class 1. The horizontal axis measures the (relative) frequency, \textit{i.e.}, the number of
    instances per million tokens.
    The different feature categories are explained in \url{https://browndw.github.io/pybiber/feature-categories.html}.}
    \label{fig:biber_features_different}
\end{figure*}

For each algorithm, after completing the training, the selected
classifier was evaluated over the test set. 
The results of that evaluation are shown in Table~\ref{tab:classifier_results}. As can be seen
in the table, all ML classifiers obtained a much higher accuracy than the accuracy
obtained by human evaluators in Section~\ref{sect:humanresults}. Moreover, two classifiers
reached an accuracy of around 0.98, with the linear classifier of~\cite{wahde_interpretable_2024}
obtaining the best result. Therefore, in the later analysis (below) that involves ML classifiers, we have employed that classifier throughout.

\section{Analysis and Discussion}
\label{sect:analysis}
The results presented above show a very large difference in performance between, on the one hand, human
evaluators and, on the other, ML classifiers. In this section, we set out to study the reasons behind
this discrepancy. First of all, the fact that human evaluators perform so poorly may be
explained by the apparent similarities between the two text categories: First, in the word frequency lists from the two classes, words such as \textit{the, and, of, to, ...} appear at the top of the list, as expected in any English language text. Second, the LLM-rewritten texts use the same proper nouns that appear in the original, human-written text. Moreover, as also shown in the linguistic feature analysis below, the two classes do not have a major distinction in the usage rates of several features such as, \textit{inter alia}, pronouns or some verb tenses, as shown in Figure~\ref{fig:biber_features_similar}. These similarities make the texts appear coherent and sensible to a human observer, making it hard to detect the underlying differences between the two text classes.

Owing to this lack of any obvious distinguishing characteristics, it is perhaps somewhat surprising to see that various text classifiers reach very high accuracies over our test data set, as shown in Table~\ref{tab:classifier_results}. However, there
are, in fact, several systematic differences between the two classes, which may be imperceptible to human evaluators
but are used with great success by the ML classifiers, as we will now demonstrate.

\subsection{Feature Analysis}
\label{sec:feature}

In~\cite{reinhart2025llms} a feature set of 67 linguistic items (lexical and grammatical), defined in~\cite{biber1991variation}, was used to classify human-written 
and LLM-generated text of various categories. The results of that study show that a distinctive use of certain features is associated with LLM-generated text. Thus, we explored whether the data set used here displays similar characteristics.
For this task, the Python package \textit{pybiber} was applied to extract the average frequencies of the features in the two text classes. An overview of some of the results is shown in Figures~\ref{fig:biber_features_similar} and~\ref{fig:biber_features_different}. The first figure displays the features where the frequencies vary the least between the two classes, thus
underlining some similarities. In the second figure, one can see the 10 features with the largest increase (left panel) and the largest decrease (right panel) in frequency in the LLM-generated texts compared to the human-written samples. In particular, this figure shows that the use of present and past participles is more prominent in the LLM-generated text.

\begin{table*}[ht]
    \caption{Detailed description of the explanation codes used in the manual annotation process (see accompanying text). The examples shown in the third column are of the form \textit{token1/token2}, where
    \textit{token1} indicates typical usage in a human-written passage, whereas \textit{token2} 
    represents the corresponding token likely used by the LLM. Note that, in some cases, the tokens represent more than one word here, but the cases can still be distinguished by considering individual words, as was done in the classifier used here.
    The final column shows the number of instances ($N$) of individual features among the 190 features that are associated with the explanation; note that some features were identified as belonging in multiple explanation categories depending on the context they appeared in; hence, the instance counts do not sum to the size of the feature subset.}
    \centering
    \begin{tabular}{llp{95mm}r}
    \hline
        Code & Description & Examples & $N$  \\
        \hline
        E1 & Rephrased text & Subcategories: &  \\
         &  & {\small{E1.1: dialogue verbs, \textit{said}/\textit{replied}, \textit{cried}/\textit{exclaimed}, \textit{laughed}/\textit{chuckled}}} & 17 \\
         &  & {\small{E1.2: other verbs, \textit{thought}/\textit{assumed}, \textit{remembered}/\textit{recalled}, \textit{think}/\textit{imagine}}} & 50 \\
         &  & {\small{E1.3: prepositions, \textit{to}/\textit{toward}}} & 5 \\
         &  & {\small{E1.4: miscellaneous, \textit{Look here}/\textit{Listen}, \textit{somebody}/\textit{someone}, \textit{But}/\textit{Yet}}} & 28 \\
         &  & {\small{E1.5: adverbs and adjectives, \textit{only}/\textit{simply}, \textit{beautiful}/\textit{exquisite}}} & 50 \\
         &  & {\small{E1.6: grammar, \textit{I thought}/\textit{I was convinced}, \textit{I am going to}/\textit{I intend to}}}  & 14 \\ 
         &  & {\small{E1.7: nouns, \textit{door}/\textit{entrance}}, \textit{body}/\textit{corpse}, \textit{prisoner}/\textit{accused}} & 33 \\
        
        E2 & Temporal drift effects & Dated spelling: \textit{to-day}/\textit{today}, 
        \textit{to-morrow}/\textit{tomorrow} & 4 \\
        E3 & English varieties &  Spelling variations (British or American): \textit{realise}/\textit{realize} \newline Word choices: \textit{I fancy}/\textit{I think}, \textit{just round the}/\textit{just around the} & 4 \\
        E4 & Foreign language & Passages in French: \textit{mon ami}/\textit{my friend} & 2 \\
        E5 & Colloquial expressions & Short form words: \textit{'em}/\textit{them} & 1 \\
        \hline
    \end{tabular}
    \label{tab:explanation_codes}
\end{table*}

While these results provide a partial explanation for the excellent performance of the ML classifiers, a more
complete analysis can be made by studying in detail the top-performing classifier, namely the linear classifier (top row in Table~\ref{tab:classifier_results}). 
Being explicitly focused on interpretability, this classifier is well suited for that type of analysis.

The analysis aimed at explaining the differences between the two classes was carried out by manually annotating the classifier features. As manual annotation is an arduous task, only a subset 
of the most frequent features was considered. Let $f_{i,j}$ denote the relative frequency of a feature $\varphi_i$ in class $j$ (where $j$ is thus either 0 or 1), measured as the number of instances (in this case, of unigrams) per million tokens.
A subset of important features was generated where each selected feature had both (1) a high relative frequency $f^{\rm{max}}_i = {\rm{max}}(f_{i,0},f_{i,1})$ in one of the classes, and (2) a high frequency ratio $R_{i}$, defined as $f^{\rm{max}}_i/(f_{i,0}+f_{i,1})$, such that
values close to 1 indicate that the feature predominantly occurs in \textit{one} of the
classes, whereas features that are more evenly used between the two classes have $R_i$
close to $0.5$.

The conditions $f^{\rm{max}}_i \ge 100$ and $R_i \ge 0.75$ were used to select features from the classifier's feature set, resulting in a subset of 190 features.
This subset was then annotated by hand, using the text sample pairs in the data set to identify the context and to study differences in the use of the studied features in the two classes. Annotation codes for the identified features are shown in Table~\ref{tab:explanation_codes}. This analysis is not meant to be comprehensive: A complete analysis would require multiple annotators and a larger subset of the features to inspect. Instead, this analysis is aimed at explaining the differences between the LLM-generated text and the human-written text from a general perspective.

\begin{figure*}[t]
    \centerline{
    \includegraphics[width=0.48\linewidth]{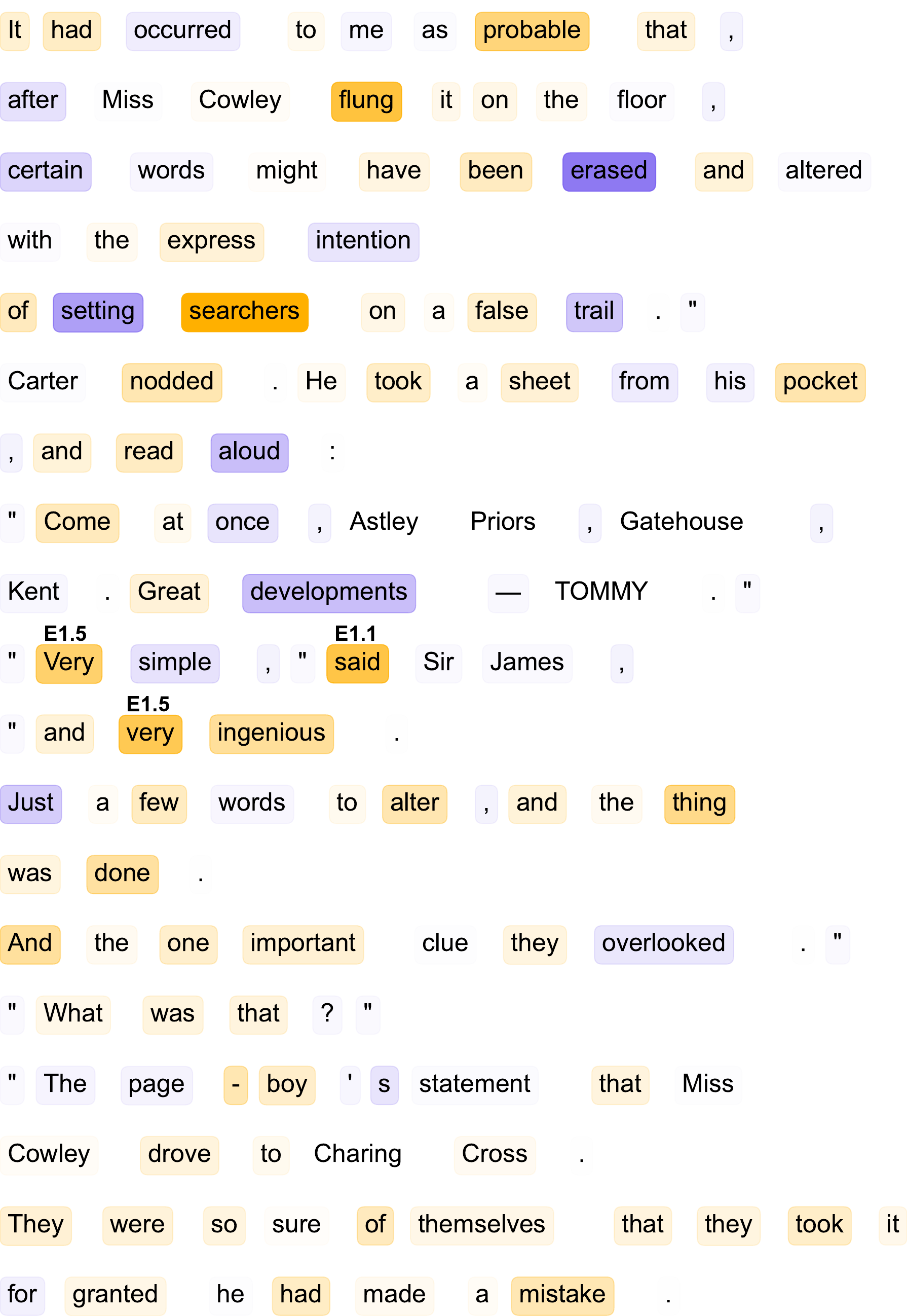}
    \includegraphics[width=0.48\linewidth]{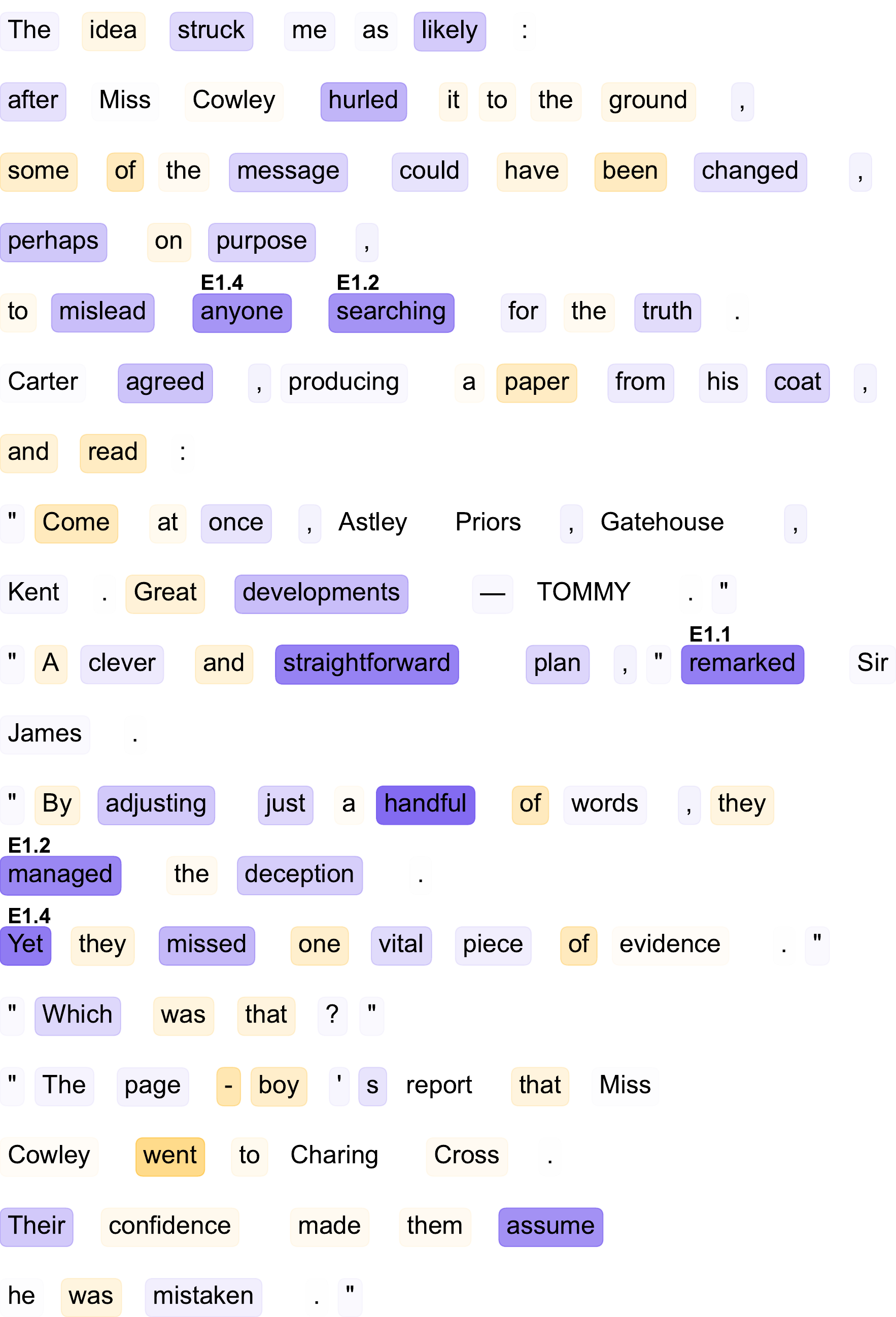}
    }
    \caption{Visualisation of the classification result using the interpretable classifier, where one can inspect the precise impact of each unigram feature on the classification result. Yellow is used to highlight negative weights (Class~0, human-written) and purple for positive weights (Class~1, LLM-generated). The text pair is a sample from the test set. Stronger colour saturations indicate higher weight magnitudes of the features. Explanation codes are shown above features that were part of the manual annotation procedure.}
    \label{fig:feature_visualization}
\end{figure*}

The most frequent explanation type (E1) in the feature subset, as seen from the instance count $N$ shown in the rightmost column of Table~\ref{tab:explanation_codes}, was related to various forms of rephrasing by the LLM. This explanation category was further split into multiple subcategories: dialogue verbs (E1.1), other verbs (E1.2), prepositions (E1.3), miscellaneous (E1.4), adverbs and adjectives (E1.5), syntactic and grammatical changes (E1.6), and nouns (E1.7). In some cases, the variations in the LLM-rewritten texts, such as the use of different prepositions or introducing adjectives into a sentence, may naturally also change the grammatical structure of the text. In addition, some words belong in multiple word classes. Thus, there may be several different explanation categories for some features. 

Focusing first on the largest explanation category, E1 (rewriting), we identified that the most common case was related to verb usage (E1.1 and E1.2). For example, we identified 17 different dialogue verbs from the feature subset, used in text to describe conversations. The vast majority of these verbs (14 verbs) were preferred by the LLM, as evidenced by their positive 
feature weights. One common case of these rewritings is the variation in the LLM-generated text when rephrasing the verb \textit{said}; the rewritings of this verb include \textit{remarked}, \textit{responded}, \textit{mentioned}, and many more. Another example is the verb \textit{cried}, often replaced by \textit{exclaimed}, which perhaps has some overlap with the temporal drift explanation (E2; see below), although not annotated as such. Besides dialogue verbs, the rewriting alters many other verbs used in the text excerpts as well. In most cases, the same verb tenses that are used in the human-written text are also used in the LLM-generated text. The main difference between the two text categories, in regards to verbs, is then the choice of verbs, that is, the LLM-generated text contains lexemes that are used less often in the human-written text.

The second-most common subcategory of rewriting concerned adverbs and adjectives (E1.5). Here, words that can be considered in some sense to be simple tend to be associated with Class 0 (human-written): 
\textit{beautiful}, \textit{bad}, \textit{good}, \textit{great}. The text generated by the LLM alters these terms and, once again, uses a larger variety of synonyms for these words. The category consisting of nouns (E1.7) also includes synonymous word replacements, as seen from the examples in Table~\ref{tab:explanation_codes}. We also identified some cases of nominalisation (see Figure~\ref{fig:biber_features_different}), for example, the noun \textit{investigation} is also used to reformulate verb phrases: \textit{I'm having the movements of the car investigated} becomes \textit{the car's route is under investigation} when formulated by the LLM.

The miscellaneous category (E1.4) contained features that were mainly associated with conversation, for example, some phrases (\textit{Look here!}), discourse markers (\textit{In fact} or \textit{Actually}), conjunctions (in particular, when used to start a sentence, such as \textit{But} or \textit{Yet}), and indefinite pronouns (\textit{somebody}, \textit{someone}).

The remaining two subcategories, namely E1.3 (prepositions) and E1.6 (grammar), were identified less often within the subset. This may be due to the limited number of samples covered in our investigation; a larger study may reveal more explanations regarding these categories. For future investigation, applying automated tools can be considered; for example, dependency parse trees could be generated and used for identifying the exact syntactic differences between the text pairs.

Other explanations seem to be less prominent within the feature subset, but nevertheless offer interesting insights, and are likely to be more frequent in the feature set as a whole. These categories include effects of temporal drift (E2), \textit{i.e}., changes in word usage over long periods of time, as well as differences in English language varieties (E3). The human-written text contains novels from the 1920s and 1930s, written by British authors, meaning that features prominent in that time period and language variety are naturally more common in the human-written text class. The LLM-rephrased text largely fails to reproduce this style of text, even though the differences are subtle and not easily detected by a human observer, as demonstrated in Section~\ref{sec:results}. In many cases, the LLM exhibits a preference for Americanisms, using words and phrases that appear very infrequently (or not at 
all) in the human-written text considered here.

Similarly, in the rewritings, any remnants of multilingual text (E4) and colloquialisms (E5) are mostly removed, perhaps deemed unnecessary or ungrammatical by the LLM, even though
the use of such features often makes a novel more vivid and interesting to read.

\begin{figure}
\includegraphics[width=\columnwidth]{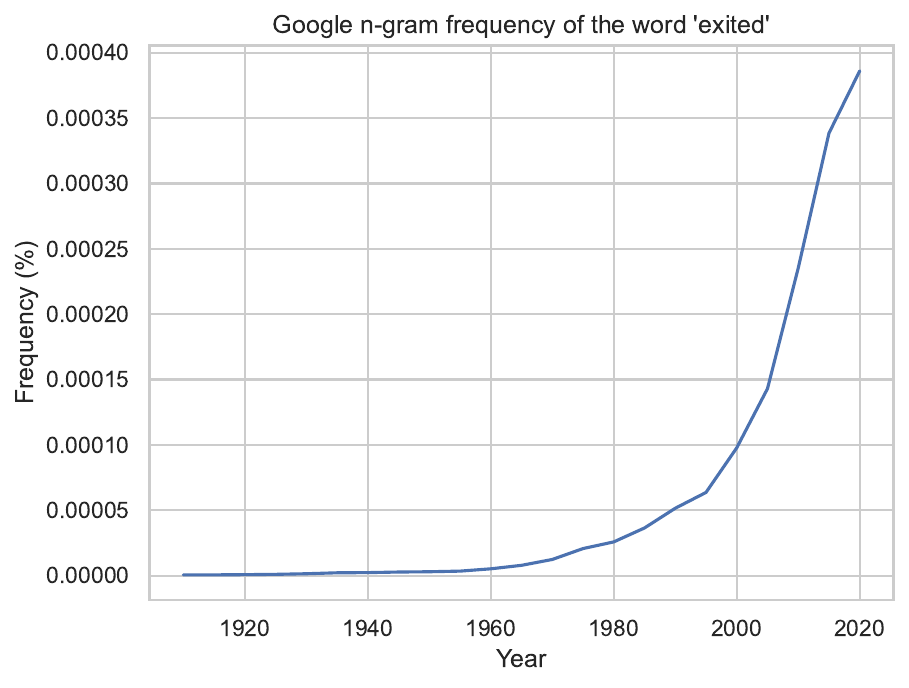}
\caption{Trend for the use of \textit{exited} from 1910 to 2020; data obtained from Google $n-$gram viewer; \url{https://books.google.com/ngrams/}.}
\label{fig:exited}
\end{figure}

A single illustrative example of temporal drift, where the frequency or meaning of a word changes over time, is shown in Figure \ref{fig:exited}. Zero-suffix derivation of
verbs from nouns is a pattern common in English, as
described in (Barbu Mititelu et al., 2023).
An example is the use of the word \textit{exit} as a denomial verb
rather than a noun. In Figure 4 we consider the past-tense
form \textit{exited} and observe that it was little
used before around 1960.
The token \textit{exited} never occurs in the source text samples (Class~0), but appears 61 times in the rewritten samples (Class~1) of
the training set, showing that the text generated by
the LLM follows contemporary language patterns atypical for the original text.

A visualisation from the classifier selected for the analysis is shown in Figure~\ref{fig:feature_visualization}. This figure shows the contribution
of each unigram feature and can be used to visually inspect why a sample was classified with a given class label: Here, the (unigram) features are highlighted in a shade of either yellow (denoting Class 0) or purple (denoting Class 1). Stronger colour saturations indicate larger feature weight magnitudes. 
Explanation codes for those features that were included in the manual annotation process are shown above the corresponding features.
Those features generally have strong feature weights also (\textit{e.g.}, \textit{very}, \textit{said}, \textit{anyone}, \textit{searching}, and so on). In the figure, one can see examples of the E1 (rewriting) category: The dialogue verb \textit{said} is replaced by \textit{remarked}, and the adverb \textit{very} is removed altogether in the LLM-generated text. Some examples of grammatical rewriting can also be identified; for example, the final expression, originally \textit{he made a mistake}, uses a reformulation that introduces the passive phrase \textit{he was mistaken}.

\begin{figure}
\centering 
\includegraphics[width=\linewidth]{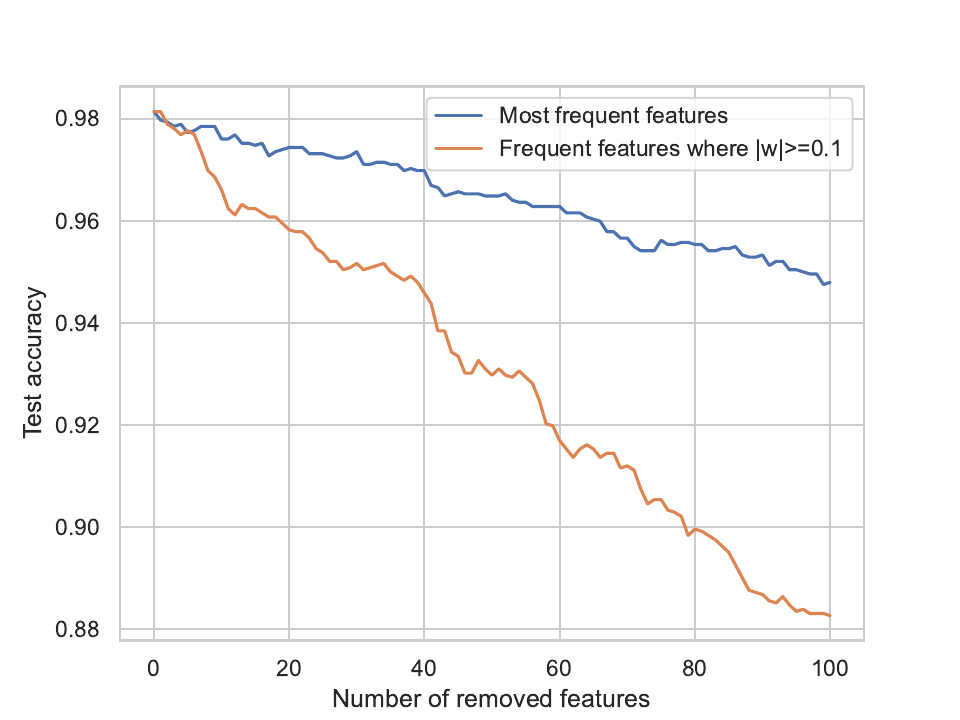}
\caption{Drop in test set accuracy when features are removed from consideration. Top (blue) curve: Removal
of features in falling order of their frequency; Bottom (orange) curve: Removal only of features
with $|w| \ge 0.1$ (again sorted in falling order of frequency). See the main text for a more
complete description.}
\label{fig:accuracydrop}
\end{figure}
Figure~\ref{fig:accuracydrop} is an attempt at showing the impact of the most salient unigrams from the perspective of classification: First, the features were sorted in
descending order based on their frequency of occurrence, here measured as $f_{i,0} + f_{i,1}$,
where $f_{i,0}$ and $f_{i,1}$ again are the relative frequencies (number of instances per million words) of
a feature in the two classes.
Then, from that list (containing all the 30,302 unigram features), the first 100 features were removed one by one 
(\textit{i.e}., their weights were set to 0, exactly). After every removal step, the classification accuracy (over the test set) was computed again,
using the remaining features, resulting in the blue curve in the figure.
Next, the procedure was repeated, again going through the same sorted list of features, but
this time only removing those features whose weight values (in the best classifier, \textit{i.e}., the
top row in Table~\ref{tab:classifier_results}) had a magnitude of at least 0.1.
After every removal, the classification accuracy was recomputed, exactly as above, producing
the orange curve in the figure. The choice of threshold (0.1) is somewhat arbitrary, but it does avoid many very common
tokens, such as \textit{the}, \textit{a}, \textit{he}, \textit{she}, and so on.

As expected, there is a drop in accuracy as these frequent features are removed. What is interesting,
however, is the \textit{magnitude} of the drop, especially when comparing the two curves: By 
removing the first 100 large-magnitude weights, one observes a drop in accuracy from 0.98 to around 0.88. However, even the lower of those values is still much better than the near-random performance of human evaluators.
The difference in the magnitude of the drop (between the two curves) is non-trivial: Even
though the weights removed in the second case (orange curve) have larger weight magnitude, they also
generally have fewer instances (since, in that case, one has to move far beyond the first 100
features to find a total of 100 features with weight magnitude of at least 0.1).
Among the large-magnitude weights removed in the second case there are many for which the
weight magnitude has no identified explanation. However, that set of removed features also includes
a few that were identified in our manual annotation process described above, for example, the features \textit{said},
\textit{little}, \textit{got}, and \textit{good}; see also Table~\ref{tab:explanation_codes}.

\subsection{Entropy Analysis}
\label{sec:entropy}
\begin{figure}[t]
\centering 
\includegraphics[width=\columnwidth]{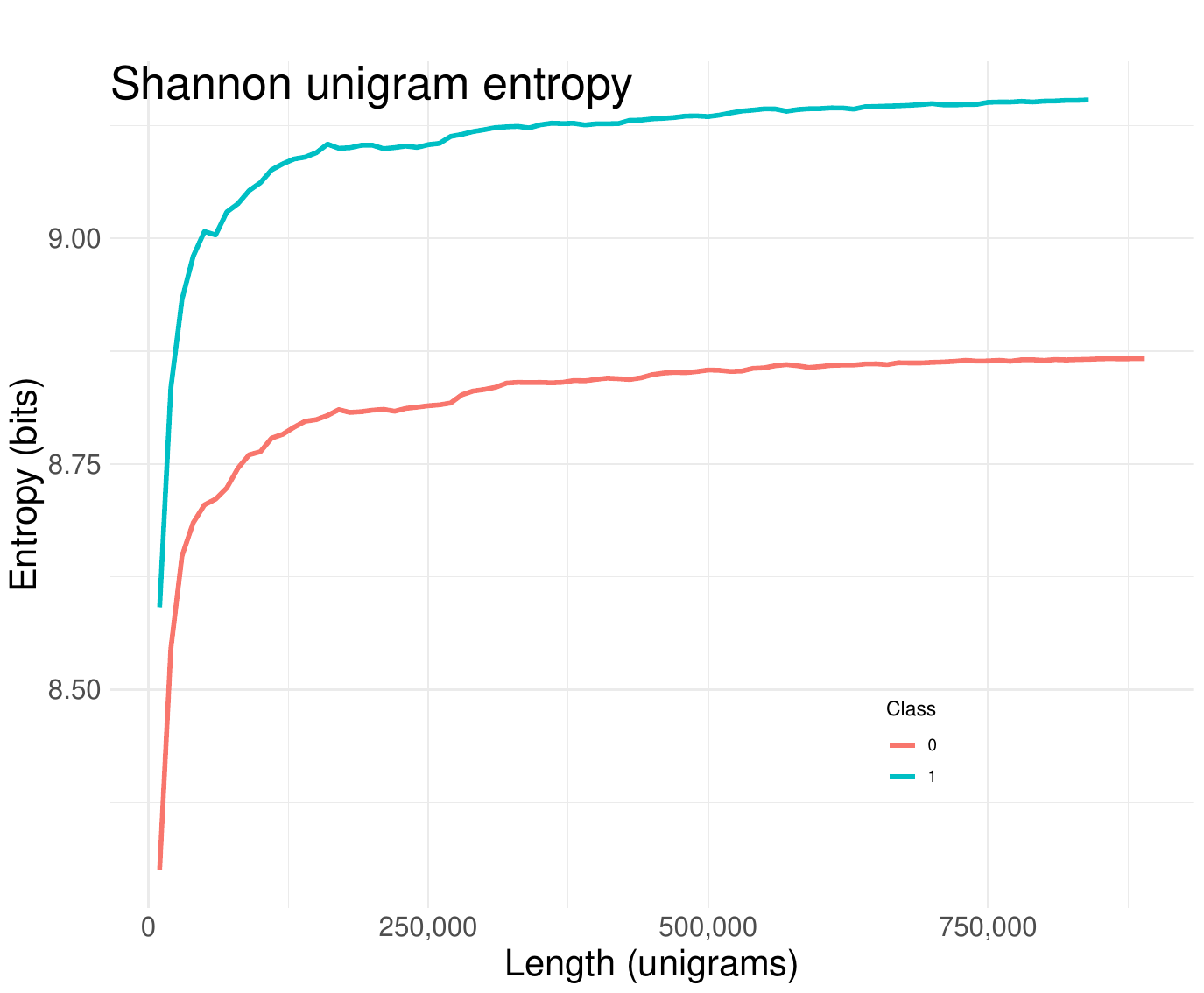}
\caption{Shannon entropy over increasing sample size.}
\label{fig:shannon}
\end{figure}

Another, more abstract, way to gain a high-level overview of the variability introduced in the rewritten text is to compare the information-theoretic entropy of the original human-written texts versus the LLM-generated texts.
Figure \ref{fig:shannon} shows $H$, the unigram entropy over different sample lengths (number of tokens) 
for Class 0 and Class 1, defined by Shannon's equation 
presented in~\cite{shannon1948}:

\begin{equation}\label{eqn:shannon}
H = -\sum_{i=1}^{n} p(t_i) \log_2 p(t_i),
\end{equation}
where $t_i$ is a unique token and $p(t_i)$ is the frequency of that token in the sample. 

We compiled all the samples from Class 0 and Class 1 into two complete sets, then
sampled the texts from each class over increasing number of tokens, taking the first 1000, 2000, and so on, computing the entropy each time. 
We see the estimated entropy levelling off at around a length of 200,000 unigrams, 
showing that we have enough data to obtain
a good estimate for the entropy of each class.
We note the higher entropy in Class 1 overall, indicating greater
variety in word choice in the rewritten texts.
The entropy over the whole data set is 0.29 bits higher for Class 1 (H =  9.164 versus H =  8.875 for Class 0). While both values are in the usual range for English text -- approximately 7 to 11 bits, with a mean of 9.14 reported in~\cite{Bentz2017Entropy} -- the higher entropy 
in Class 1 indicates greater variation in the vocabulary 
used compared to the original human-written texts,
corroborating our hypothesis that a greater range of
vocabulary is introduced by the rewriting process, as shown in the
feature analysis.
As $2^{9.164}/2^{8.875} \approx 1.22$, we expect that the rewritten texts will show around 22\% greater variation in their word-choice.

\section{\uppercase{Conclusions and Further Work}}
\label{sec:conclusion}

We have confirmed and extended earlier work showing that simple linear classifiers can 
reliably distinguish human-written text from LLM-generated text in the style
of classic detective novels, whereas human observers generally cannot. 
Our detailed analysis, and the accompanying visualisation (see Figure~\ref{fig:feature_visualization}), shows that the classification depends on
a constellation of features, principally greater variability in word choice, 
influenced by temporal drift in the usage of certain words, and a tendency by the LLM to avoid colloquialisms and to introduce
Americanisms in the generated text. 

One limitation of the study is that the methodology may have introduced greater 
variation in the LLM-generated texts by requesting rewording; such an approach may
invite use of synonyms, for example. Alternative approaches would include prompting the 
LLM to write a scene from a brief outline, with no sample text provided; or to 
present one or more samples of human-written text, and prompt for a continuation.
Some preliminary experiments suggest that the generated text remains highly 
detectable using these methods, despite removal of the explicit rewriting process.
However, further work is required to confirm this result and to establish the effects
of different generation methods overall.
Other lines of future investigation include analysing text produced by different LLMs,
in various genres, and with different prompts.

The interpretability of the results
give confidence that the classifiers are robust, while sidestepping our 
\textit{paradox of interpretability}: The interpretation presented here does not give malicious actors any \textit{simple} recipe for generating more deceptive text. 
Indeed, Figure
\ref{fig:accuracydrop} implies that it is difficult to make automatically generated
detective fiction harder to detect by applying post-processing to obviate characteristic
features, whether by removal or by rewriting. As shown in Fig.~\ref{fig:accuracydrop}, if such post-processing could
nullify the effect of the 80 features which most help the classifiers 
-- a daunting level of post-processing -- the linear classifier would still have an accuracy of around 90\%. It should be possible therefore to make tools that integrate with the workflow of editors or publishers to give a reliable indication of LLM-generated text, 
mitigating attempts to misrepresent it as human work, and thereby helping to protect
the creative contribution and livelihood of creative authors.

\bibliographystyle{apalike}
{\small
\bibliography{main}}

\end{document}